\pdfoutput=1

\documentclass[11pt]{article}

\usepackage[final]{acl}

\usepackage{float}
\usepackage{times}
\usepackage{latexsym}
\usepackage{tcolorbox}
\usepackage{xcolor}
\usepackage{ulem}
\usepackage{tikz}
\usepackage{caption}
\usepackage{hyperref}
\hypersetup{pdfauthor={Anonymous}}

\definecolor{custompink}{HTML}{EC69BA}
\definecolor{customblue}{HTML}{6BBAEA}
\definecolor{customgreen}{HTML}{86c738}

\usepackage[T1]{fontenc}

\usepackage[utf8]{inputenc}

\usepackage{microtype}

\usepackage{inconsolata}

\usepackage{graphicx}

\usepackage{booktabs}

\usepackage{titlesec}
\usepackage{enumitem}

\title{Does ``Reasoning'' with Large Language Models Improve Recognizing, Generating and Reframing Unhelpful Thoughts?}

\author{\parbox{0.9\linewidth}{
\centering{Yilin Qi\thanks{Equal Contribution.}$^{\heartsuit}$ ~~~ Dong Won Lee\footnotemark[1]$^{\spadesuit}$ ~~~ Cynthia Breazeal$^{\spadesuit}$ ~~~ Hae Won Park$^{\spadesuit}$
} \\
{\rm $^\heartsuit$Harvard University~~$^\spadesuit$MIT} \\
}
}

\begin{document}
\maketitle
\begin{abstract}
Cognitive Reframing, a core element of Cognitive Behavioral Therapy (CBT), helps individuals reinterpret negative experiences by finding positive meaning. Recent advances in Large Language Models (LLMs) have demonstrated improved performance through reasoning-based strategies. This inspires a promising direction of leveraging the reasoning capabilities of LLMs to improve CBT and mental reframing by simulating the process of critical thinking, potentially enabling more effective recognition, generation and reframing of cognitive distortions. In this work, we investigate the role of various reasoning methods, including pre-trained reasoning LLMs, such as DeepSeek-R1, and augmented reasoning strategies, such as CoT \cite{wei2022chain} and self-consistency \cite{wang2022self}, in enhancing LLMs' ability to perform cognitive reframing tasks. We find that augmented reasoning methods, even when applied to ``outdated'' LLMs like GPT-3.5, consistently outperform state-of-the-art pretrained reasoning models such as DeepSeek-R1 \cite{guo2025deepseek} and o1 \cite{jaech2024openai} on recognizing, generating and reframing unhelpful thoughts. 
\end{abstract}

\section{Introduction}
Cognitive Behavioral Therapy (CBT) \cite{beck1963thinking} is one of the most widely used and well-supported approaches in psychotherapy \cite{fenn2013key}. CBT focuses on both the process and content of thoughts, including core beliefs, assumptions, and automatic thoughts \cite{fenn2013key}. Cognitive Reframing is central to CBT, helping individuals reinterpret negative experiences by critically reasoning through and aligning them with their belief systems to find purpose or positive meaning in adversity \cite{BLUM2012596}. Recent advancement in Large Language Models (LLMs) research have focused on reasoning, which stands out as a fundamental element of human intelligence that drives key processes like problem-solving, decision-making, and critical thinking \cite{huang2022towards}. Furthermore, LLMs that incorporate reasoning in its pretraining phase or as a post-hoc augmentation procedure have shown significant improvement in performance across many tasks \cite{qiao2022reasoning}. 

In this paper, we investigate the extent to which reasoning can improve LLM's ability in Cognitive Reframing. We implement and evaluate three conditions of LLM reasoning on established cognitive reframing tasks, which include generating, recognizing, and reframing unhelpful thoughts. In addition, we propose a novel task of reframing thoughts conditioned on reframing strategies based on positive psychology \cite{harris2007integrating}.  The reasoning conditions we evaluate include: (1) LLMs pre-trained specifically for reasoning; (2) LLMs augmented with state-of-the-art reasoning methods such as CoT \cite{wei2022chain}, ToT \cite{yao2023tree}, and self-consistency \cite{wang2022self} and DoT \cite{chen2023empowering}; and (3) Non-reasoning LLMs that were not explicitly trained or augmented with reasoning capabilities. We find that reasoning-augmented models consistently outperform pre-trained reasoning models, suggesting that simply augmenting LLMs with reasoning strategies can provide strong performance gains on cognitive reframing tasks without the cost and complexity of pretraining explicitly for reasoning.

\section{Related Work}
\label{related work}
\noindent \textbf{Early AI Systems for Cognitive Reframing}
Early mental health chatbots and apps incorporated elements of Cognitive Reframing, but relied on scripted responses or simple AI \cite{hodson2024can}. Systems like the CBT-based chatbot Wysa could walk users through CBT-style prompts by using AI to select from pre-written therapist responses, but they lacked the flexibility to produce personalized new reframes \cite{hodson2024can}.

\noindent \textbf{LLMs for Identifying and Reframing Unhelpful Thoughts}
Recent studies have begun leveraging LLMs to identify and reframe unhelpful thoughts in more flexible ways. Previous work explored LLM-assisted cognitive reframing by training a retrieval-augmented model to suggest alternative thoughts with controlled therapeutic attributes \cite{sharma2023cognitive}. Others introduced a structured “Diagnosis of Thought” prompting technique that guides the model to separate facts from subjective interpretations and reason about evidence, significantly improving the detection of distorted thinking patterns while producing expert-approved explanatory rationales \cite{chen2023empowering}. These works demonstrate the feasibility of LLMs both in generating helpful reframed thoughts and in pinpointing unhelpful thinking. 

\noindent \textbf{Therapeutic Frameworks and Prompt Engineering}
To further enhance LLM-based cognitive restructuring, researchers have applied explicit therapeutic frameworks and structured prompting. RESORT framework provides a series of psychologically grounded reappraisal instructions \cite{zhan2024large}. Similarly, the HealMe system integrated core CBT techniques into the prompt structure, systematically guiding the LLM to distinguish circumstances from feelings, brainstorm alternative perspectives, and develop empathetic, actionable new thoughts \cite{xiao2024healme}.

\section{Experiments}
\label{experiments}
In this work, we investigate the contribution of reasoning methods in cognitive reframing. We utilize the PatternReframe dataset \cite{maddela-etal-2023-training}, where each sample contains (1) a persona (i.e \textit{"I enjoy gardening. My favorite drink is red wine. I work for a clothes retailer. I have one child."}), (2) unhelpful thought (i.e. \textit{"My child wishes they had another sibling. I bet they think I'm a horrible parent for stopping at one child."}), (3) the unhelpful thinking pattern (i.e. \textit{"Jumping to conclusions: mind reading"}), and (4) the reframed positive thought (i.e. \textit{"My child wishes they had another sibling, but I'm grateful I can focus all my attention on one child."}) and the aligned reframe strategy (i.e. \textit{"Optimism"}). The unhelpful thinking patterns as well as strategies used to reframe unhelpful thoughts are both grounded in psychology literature \cite{david1980feeling}, \cite{harris2007integrating}. We sample a set of 1,000 examples from the dataset such that the occurence of each unhelpful thinking pattern is distributed uniformly ($\sim$100 per category, e.g., Personalization, Catastrophizing) for use across all tasks.

\noindent \textbf{Methods} We experiment with three broad conditions of LLM models and reasoning methods. \textbf{(1) \emph{N}on-\emph{R}easoning (\emph{NR})} models include those that have not been specifically trained for reasoning purposes. In our experiments, we focus on \texttt{GPT-3.5}, \texttt{GPT-4}, \texttt{GPT-4o}. On the other hand, we also consider \textbf{(2) \emph{P}retrained \emph{R}easoning (\emph{PR})} models that have been specifically trained for reasoning, these include \texttt{Llama-3.3}, \texttt{Deepseek-R1} \cite{guo2025deepseek}, \texttt{GPT-o1} and \texttt{GPT-o3-mini}. Finally, to study the effects of modern reasoning methods and prevent confounding analysis due to data leakage, we utilize \texttt{GPT-3.5} as the base model, as other recent models' data cutoff date is beyond the data release date for PatternReframe (Jul 2023). We consider popular state-of-the-art \textbf{(3) \emph{A}ugmented \emph{R}easoning (\emph{AR})} methods described below: 

\noindent\textbf{Chain-of-Thought (CoT) \cite{kojima2022large,wei2022chain}}: supplies LLMs with step-by-step reasoning demonstrations instead of conventional input-output pairs. We focus on the popular technique of \texttt{zero-shot CoT}, where a simple prompt of ``Let’s think step by step`` is prepended to the prompt to facilitate step-by-step thinking. 

\noindent\textbf{Self-Consistency (SC) \cite{wang2022self}}: is a reasoning method based on the decoding strategy, self-consistency. Instead of selecting a single greedy path, it samples a diverse set of reasoning paths and determines the most consistent answer by marginalizing over these sampled paths. 

\noindent\textbf{Tree-of-Thought (ToT) \cite{yao2023tree}}: is a framework that enhances language models' problem-solving by exploring multiple reasoning paths structured as a tree. Each node represents a partial solution, and the model generates, evaluates, and searches through these "thoughts" using strategies like breadth-first (BFS) or depth-first search (DFS). In our experiments, we use DFS. 

\noindent\textbf{Diagnosis-of-Thought (DoT) \cite{chen2023empowering}}: is the most relevant to our work and was previously proposed for the same task of cognitive distortion detection. The method diagnoses a patient's speech through three stages: subjectivity assessment to distinguish facts from thoughts, contrastive reasoning to explore reasoning processes that support or contradict the thoughts, and schema analysis to summarize cognitive schemas.

\begin{figure}[ht]
    \centering
    \includegraphics[width=0.9\linewidth]{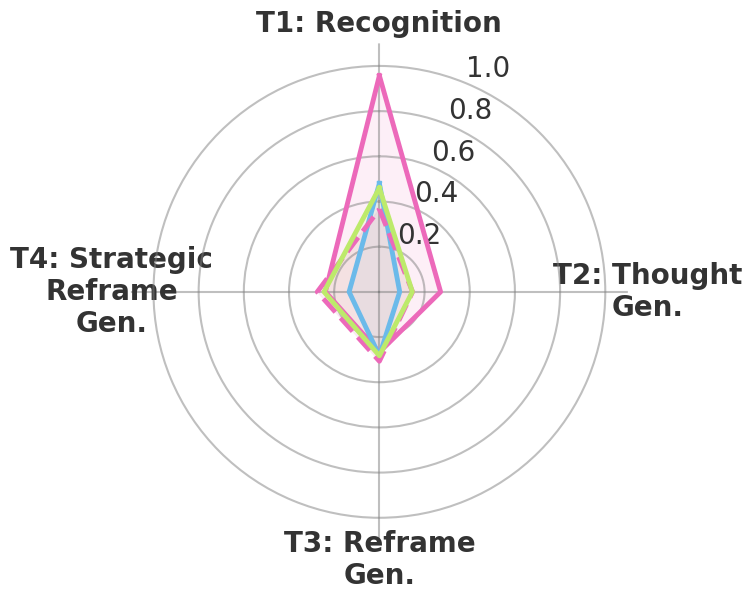}
    \caption{Performance for Representative Models in Each Class of Reasoning. 
    \textcolor{customblue}{Non-Reasoning Method \rule[0.5ex]{0.4cm}{0.4mm}}{\hspace{0.2em}}: \texttt{GPT-4o}; 
    \textcolor{customgreen}{Pre-trained Reasoning Method \rule[0.5ex]{0.4cm}{0.4mm}}{\hspace{0.2em}}: \texttt{o1};  
    \textcolor{custompink}{Reasoning-Augmented Method \rule[0.5ex]{0.4cm}{0.4mm}}{\hspace{0.2em}}: \texttt{GPT-3.5 + DoT};  
    \textcolor{custompink}{\rule[0.5ex]{0.15cm}{0.4mm}}{\hspace{0.2em}}\textcolor{custompink}{\rule[0.5ex]{0.15cm}{0.4mm}}{\hspace{0.2em}}: \texttt{GPT-3.5 + Self-Consistency}.}
    \label{fig:radar}
    \vspace{-5mm}
\end{figure}

\section{Tasks \& Results}
\label{tasks}
To evaluate the effectiveness of varying conditions of modern LLM reasoning  methods, we incorporate the following tasks: (1) recognizing unhelpful thought patterns, (2) generating unhelpful thoughts, and (3) generating reframes of unhelpful thoughts, in line with the proposed tasks from PatternReframe \cite{maddela-etal-2023-training}. Given the advances of instruction tuning and alignment \cite{ouyang2022training}, we propose a novel (4)-th task: generating strategic reframes of unhelpful thought, strictly enforcing the reframe of the unhelpful thought to be \emph{aligned} to a specific reframing strategy. The performance of representative models from each condition (PR, NR, AR) are shown in Fig. \ref{fig:radar}, where we find that simple augmented reasoning methods perform well across all tasks, and obtain massive performance gains for the task of unhelpful thought pattern recogntion.

\begin{figure}[ht]
    \centering
    \includegraphics[width=1\linewidth]{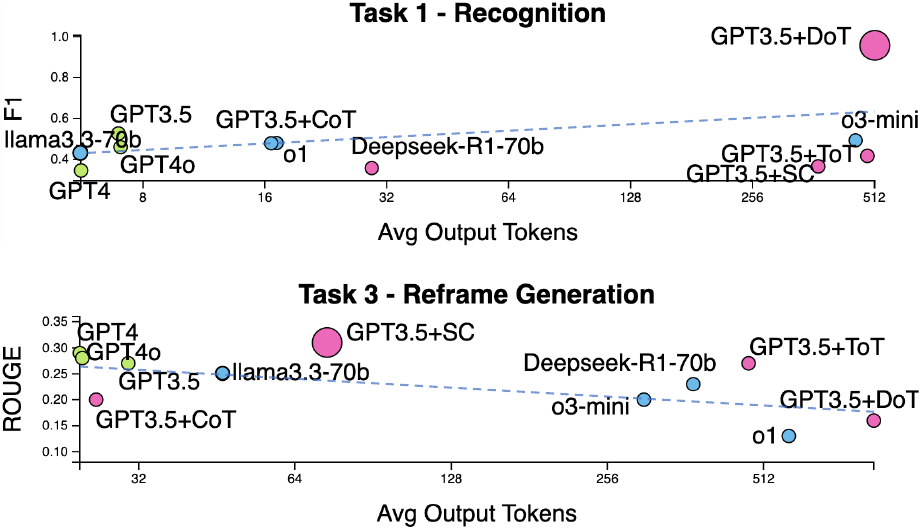}
    \caption{
        Output Tokens compared to Performance for each method across Tasks 1,3 
        (\textcolor{custompink}{\raisebox{-0.75ex}{\scalebox{2}{\textbullet}}}: Reasoning-Augmented models; 
        \textcolor{customblue}{\raisebox{-0.75ex}{\scalebox{2}{\textbullet}}}: Pretrained reasoning models; 
        \textcolor{customgreen}{\raisebox{-0.75ex}{\scalebox{2}{\textbullet}}}: Non-Reasoning models). 
        As indicated by the best performing model, encoded with a larger circle, we find that Reasoning-Augmented models can outperform Pretrained reasoning models. 
        \textcolor{customblue}{\rule[0.5ex]{0.15cm}{0.4mm}}\hspace{0.2em}
        \textcolor{customblue}{\rule[0.5ex]{0.15cm}{0.4mm}}\hspace{0.2em}: Linear Regression fit on average output tokens to performance.  
        We see a positive linear relationship between number of output tokens and performance for the task of recognition and a negative relationship for reframe generation.
    }
    \label{fig:metrics}
\end{figure}

\paragraph{Task 1: Recognition of Unhelpful Thought Patterns}assesses whether LLMs can recognize the unhelpful thinking pattern given a description of the persona and the unhelpful thought. An example prompt for this task can be found in App. \ref{app:task1}. We conduct an automatic performance evaluation using F1-score, accuracy, precision, and recall from prior literature \cite{maddela-etal-2023-training}. The results for Task 1 are presented in Table \ref{tab:task1}. While pretrained reasoning (PR) methods generally outperform non-reasoning (NR) methods, a simple augmentation of the \texttt{GPT-3.5} model with \texttt{DoT} (AR) achieves a remarkable performance across all metrics, outperforming the strongest pre-trained reasoning models, i.e. DeepSeek-R1 and o1, by a big margin of $\sim$ 40\% in accuracy scores. Notably, \texttt{DoT} is specifically tailored for the task of cognitive distortion detection, which aligns directly with the set-up of Task 1. \emph{These results imply that, in recognizing unhelpful thought patterns, minimally adapting LLMs with task-aligned augmented reasoning methods can significantly surpass the performance of general-purpose reasoning models.} However, while not requiring extensive fine-tuning, AR methods like \texttt{DoT} are the most computationally expensive, as reflected by their high token usage (see Fig. \ref{fig:metrics}).

\begin{table}[ht]
\centering
\resizebox{\linewidth}{!}{%
\begin{tabular}{@{}lcccc@{}}
\toprule
\textbf{Model} & \textbf{Acc.} & \textbf{Precision} & \textbf{Recall} & \textbf{F1} \\ \midrule
(NR)  GPT-3.5  & 0.425 $\pm$ 0.037 & 0.457 $\pm$ 0.055 & 0.362 $\pm$ 0.034 & 0.346 $\pm$ 0.048 \\
(NR) GPT4  & 0.504 $\pm$ 0.018 & 0.529 $\pm$ 0.024 & 0.459 $\pm$ 0.005 & 0.435 $\pm$ 0.021 \\
(NR) GPT4o  & 0.597 $\pm$ 0.037 & 0.532 $\pm$ 0.034 & 0.478 $\pm$ 0.014 & 0.460 $\pm$ 0.028 \\ \midrule
(PR) Llama-3.3  & 0.558 $\pm$ 0.025 & 0.556 $\pm$ 0.034 & 0.528 $\pm$ 0.032 & 0.527 $\pm$ 0.039 \\
(PR) o1  & 0.560 $\pm$ 0.040 & 0.550 $\pm$ 0.048 & 0.490 $\pm$ 0.020 & 0.480 $\pm$ 0.036 \\
(PR) o3-mini  & 0.549 $\pm$ 0.029 & 0.558 $\pm$ 0.054 & 0.510 $\pm$ 0.046 & 0.493 $\pm$ 0.047 \\
(PR) Deepseek-R1-70B  & 0.527 $\pm$ 0.047 & 0.522 $\pm$ 0.041 & 0.480 $\pm$ 0.037 & 0.479 $\pm$ 0.041 \\ \midrule
(AR) GPT3.5 + CoT  & 0.395 $\pm$ 0.052 & 0.41 $\pm$ 0.057 & 0.391 $\pm$ 0.040 & 0.358 $\pm$ 0.053 \\
\textbf{(AR) GPT3.5 + DoT}  & \textbf{0.956 ± 0.011} & \textbf{0.959 ± 0.011} & \textbf{0.959 ± 0.008} & \textbf{0.957 ± 0.011} \\
(AR) GPT3.5 + SC  & 0.419 $\pm$ 0.036 & 0.479 $\pm$ 0.028 & 0.371 $\pm$ 0.023 & 0.366 $\pm$ 0.027 \\
(AR) GPT3.5 + ToT & 0.434 $\pm$ 0.018 & 0.515 $\pm$ 0.050 & 0.415 $\pm$ 0.025 & 0.417 + 0.028 \\ \bottomrule
\end{tabular}%
}
\caption{Accuracy, Precision, Recall, F1 Scores for Task 1: Recognition of Unhelpful Thought Patterns.}
\vspace{-5mm}

\label{tab:task1}
\end{table} 
\begin{table}[ht]
\centering
\resizebox{\linewidth}{!}{%
\begin{tabular}{@{}llll@{}}
\toprule
\textbf{Model} & \multicolumn{1}{c}{\textbf{ROUGE}} & \multicolumn{1}{c}{\textbf{BScore}} & \multicolumn{1}{c}{\textbf{mE5 Sim.}} \\ \midrule
(NR) GPT-3.5 & 0.150 ± 0.084 & 0.874 ± 0.017 & 0.842 ± 0.039 \\
(NR) GPT4 & 0.145 + 0.093 & 0.876 + 0.018 & 0.844 + 0.040 \\
(NR) GPT4o & 0.146 + 0.091 & 0.876 + 0.018 & 0.845 + 0.039 \\ \midrule
(PR) Llama-3.3 & 0.139 + 0.064 & 0.867 + 0.015 & 0.851 + 0.034 \\
(PR) o1 & 0.090 + 0.070 & 0.823 + 0.191 & 0.850 + 0.030 \\
(PR) o3-mini & 0.121 + 0.057 & 0.858 + 0.013 & 0.850 + 0.027 \\
(PR) Deepseek-R1-70B & 0.142 + 0.081 & 0.873 + 0.017 & 0.841 + 0.038 \\ \midrule
(AR) GPT3.5 + CoT & 0.147 ± 0.085 & 0.872 ± 0.017 & 0.843 ± 0.038 \\
\textbf{(AR) GPT3.5 + DoT}
 & \textbf{0.271 ± 0.186} & \textbf{0.899 ± 0.031} & \textbf{0.884 ± 0.052} \\
(AR) GPT3.5 + SC & 0.147 ± 0.085 & 0.874 ± 0.017 & 0.844 ± 0.039 \\
(AR) GPT3.5 + ToT & 0.146 + 0.085 & 0.873 + 0.017 & 0.841 + 0.042 \\ \bottomrule
\end{tabular}%
}
\caption{Task 2 -- ROUGE, BertScore, mE5 embedding similarity scores for Generation of Unhelpful Thought}
\vspace{-5mm}
\label{tab:task2}
\end{table}
\paragraph {Task 2: Generation of Unhelpful Thought}assesses how well LLMs can generate an unhelpful thought given a persona and unhelpful thought pattern as shown in App. \ref{app:task2}. For automatic performance evaluation on this task, we report the ROUGE \cite{lin2004rouge}, BERTScore \cite{zhang2019bertscore}, and a sentence similarity metric using the \texttt{multilingual-e5-large-instruct} embedding model \cite{wang2024multilingual} – one of the top-5 best performing embedding models for retrieval on the MTEB benchmark \cite{enevoldsen2025mmtebmassivemultilingualtext}. As seen in Table \ref{tab:task2}, the non-reasoning \texttt{GPT-3.5} model augmented with \texttt{DoT} (AR) again emerges as the best-performing variant across all metrics, outperforming the strongest pre-trained reasoning model \texttt{Deepseek-R1} by 0.138 in ROUGE score. To further clarify, \texttt{DoT} is specifically designed for the detecting cognitive distortion types, not the generation of unhelpful thought. This surprising result extends the findings from Task 1, reinforcing the idea that \emph{task-related reasoning strategies not only outperform general pretrained reasoning models but can also generalize well to adjacent tasks within the same domain.} 

\paragraph{Task 3: Reframing of Unhelpful Thought} is used to assess how well LLMs can generate a reframe of the persona’s unhelpful thought given a persona, an unhelpful thought, and the unhelpful thinking pattern. An example is shown in App. \ref{app:task3}. As displayed in Table \ref{tab:task3}, we find that augmented reasoning (AR) methods again outperform all pretrained reasoning (PR) and non-reasoning (NR) methods. Specifically, \texttt{GPT-3.5} augmented with \texttt{Self-Consistency} is the best-performing variant for the task of Reframe Generation. This may be attributed to the nature of the task, which likely benefits from exploring diverse reasoning paths to produce varied yet coherent reframes. Moreover, this AR method offers a noticeable reduction in computational cost compared to other high-performing variants (see Fig. \ref{fig:metrics}), making it an effective and efficient choice for this task. The \texttt{Self-Consistency}-augmented \texttt{GPT-3.5} model exhibits this favorable trend across Tasks 2, 3, and 4 (see App. \ref{app:tokens}).

\begin{table}[ht]
\centering
\resizebox{\linewidth}{!}{%
\begin{tabular}{@{}llll@{}}
\toprule
\textbf{Model} & \multicolumn{1}{c}{\textbf{ROUGE}} & \multicolumn{1}{c}{\textbf{BScore}} & \multicolumn{1}{c}{\textbf{mE5 Sim.}} \\ \midrule
(NR) GPT-3.5 & 0.287 + 0.130 & 0.904 + 0.020 & 0.902 + 0.032 \\
(NR) GPT4 & 0.270 + 0.119 & 0.900 + 0.019 & 0.906 + 0.02 \\
(NR) GPT4o & 0.283 + 0.136 & 0.904 + 0.021 & 0.904 + 0.032 \\ \midrule
(PR) Llama-3.3 & 0.247 + 0.102 & 0.895 + 0.017 & 0.901 + 0.031 \\
(PR) o1 & 0.126 + 0.042 & 0.865 + 0.136 & 0.886 + 0.033 \\
(PR) o3-mini & 0.203 + 0.087 & 0.888 + 0.016 & 0.890 + 0.030 \\
(PR) Deepseek-R1-70B & 0.228 + 0.102 & 0.894 + 0.019 & 0.897 + 0.032 \\ \midrule
(AR) GPT3.5 + CoT & 0.196 + 0.121 & 0.885 + 0.023 & 0.872 + 0.050 \\
(AR) GPT3.5 + DoT & 0.267 + 0.126 & 0.899 + 0.019 & 0.898 + 0.032 \\
\textbf{(AR) GPT3.5 + SC} & \textbf{0.307 + 0.135} & \textbf{0.907 + 0.019} & \textbf{0.906 + 0.032} \\
(AR) GPT3.5 + ToT & 0.160 + 0.099 & 0.870 + 0.024 & 0.859 + 0.046 \\ \bottomrule
\end{tabular}%
}
\caption{Task 3: Reframing of Unhelpful Thought }
\vspace{-3mm}
\label{tab:task3}
\end{table}
\begin{table}[ht]
\centering
\resizebox{\linewidth}{!}{%
\begin{tabular}{@{}llll@{}}
\toprule
\textbf{Model} & \multicolumn{1}{c}{\textbf{ROUGE}} & \multicolumn{1}{c}{\textbf{BScore}} & \multicolumn{1}{c}{\textbf{mE5 Sim.}} \\ \midrule
\textbf{(NR) GPT-3.5} & \textbf{0.272 + 0.129} & \textbf{0.902 + 0.019} & \textbf{0.901 + 0.032} \\
(NR) GPT4 & 0.238 + 0.105 & 0.895 + 0.018 & 0.902 + 0.029 \\
(NR) GPT4o & 0.245 + 0.124 & 0.897 + 0.019 & 0.900 + 0.032 \\ \midrule
(PR) Llama-3.3 & 0.208 + 0.087 & 0.887 + 0.016 & 0.895 + 0.029 \\
(PR) o1 & 0.134 + 0.031 & 0.825 + 0.173 & 0.809 + 0.038 \\
(PR) o3-mini & 0.184 + 0.082 & 0.884 + 0.015 & 0.886 + 0.030 \\
(PR) Deepseek-R1-70B & 0.203 + 0.091 & 0.888 + 0.017 & 0.892 + 0.031 \\ \midrule
(AR) GPT3.5 + CoT & 0.200 + 0.112 & 0.888 + 0.019 & 0.881 + 0.040 \\
(AR) GPT3.5 + DoT & 0.239 + 0.106 & 0.895 + 0.018 & 0.895 + 0.031 \\
\textbf{(AR) GPT3.5 + SC} & \textbf{0.275 + 0.127} & \textbf{0.903 + 0.020} & \textbf{0.903 + 0.031} \\
(AR) GPT3.5 + ToT & 0.166 + 0.109 & 0.870 + 0.029 & 0.854 + 0.046 \\ \bottomrule
\end{tabular}%
}
\caption{Task 4: Strategic Reframing of Unhelpful Thought}
\vspace{-5mm}
\label{tab:task4}
\end{table}
\paragraph{Task 4: Strategic Reframing of Unhelpful Thought} 
We introduce a novel task that extends Task 3, aiming to evaluate how effectively large language models (LLMs) can generate a reframe of the persona’s unhelpful thought \textit{aligned to a specific reframe strategy} \cite{harris2007integrating}. This task specifically measures the alignment and instruction-tuning capabilities of LLMs in Cognitive Reframing, which is particularly important in CBT practices, where the intervention used is chosen and tailored to the specific formulation of the individual \cite{fenn2013key}. An example of the task implementation is shown in App. \ref{app:task4}. The results for Task 4 are shown in Table \ref{tab:task4}. Surprisingly, we find that the non-reasoning (NR) version of \texttt{GPT-3.5} and its \texttt{Self-Consistency}-augmented (AR) variant display the strongest but similar  performance over other methods. In addition, overall performance on Task 4 \ref{tab:task4} is lower than Task 3 \ref{tab:task3}. These two results combined indicate that even the most advanced pretrained and augmented reasoning (PR, AR) models lack sufficient alignment to be able to generate mental reframes that are strictly aligned to specific reframe strategies. These findings underscore the need for further research on developing alignment and controllable generation methods for LLMs to be effectively and reliably used for CBT applications.

\bibliography{custom}

\clearpage

\appendix 
\label{sec:appendix}
\section{Relationship Between Output Tokens and Performance}
\label{app:tokens}
\begin{figure}[H]
    \centering
    \includegraphics[width=1\linewidth]{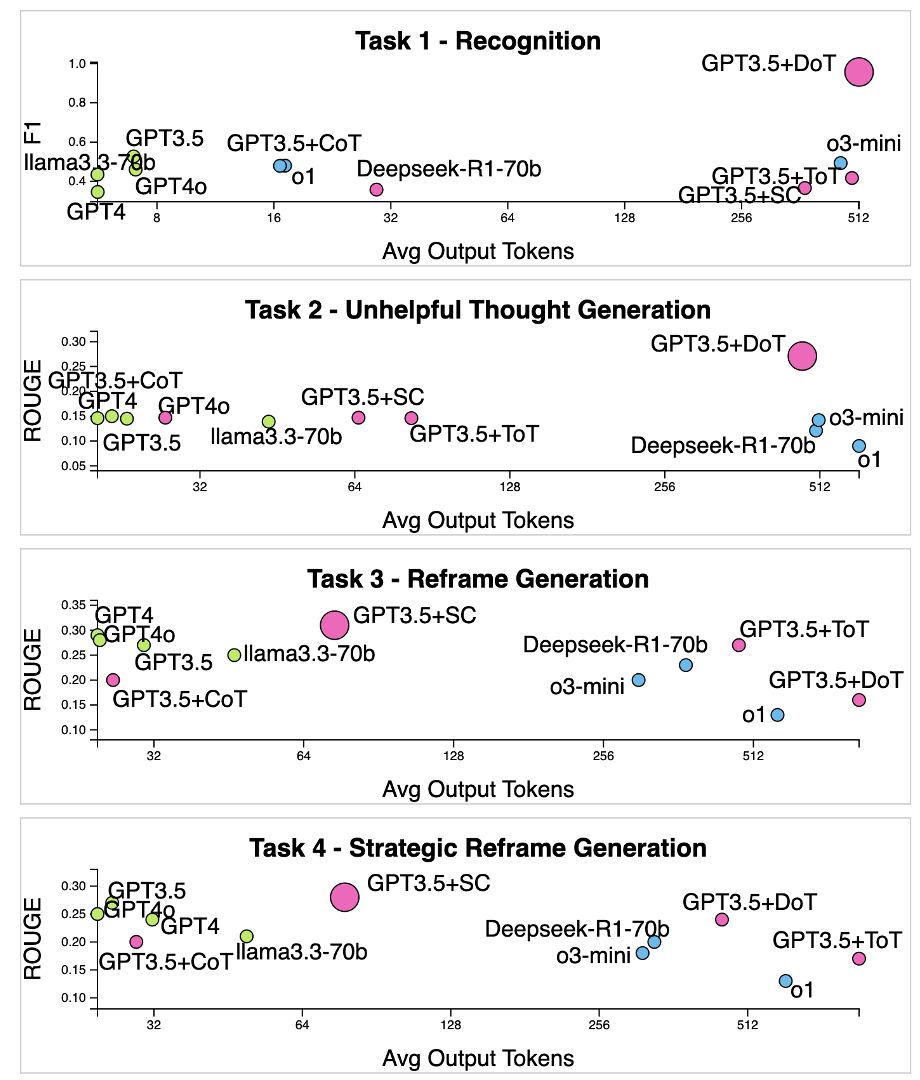}
    \caption{
        Output Tokens compared to Performance for each method across Tasks 1, 2, 3, 4 
        (\textcolor{custompink}{\raisebox{-0.75ex}{\scalebox{2}{\textbullet}}}: Reasoning-Augmented models; 
        \textcolor{customblue}{\raisebox{-0.75ex}{\scalebox{2}{\textbullet}}}: Pretrained reasoning models; 
        \textcolor{customgreen}{\raisebox{-0.75ex}{\scalebox{2}{\textbullet}}}: Non-Reasoning models). 
        As indicated by the best performing model, encoded with a larger circle, we find that Reasoning-Augmented models can outperform Pretrained reasoning models. 
    }
    \label{fig:app-metrics}
\end{figure}

\section{Prompts Used}
\label{app:prompts}

The reframing strategy Definitions:

\begin{itemize}[nosep, leftmargin=*]
    \item "Growth Mindset": Reframe a challenging event as an opportunity to grow instead of dwelling on the setbacks.
    \item "Impermanence": Say that bad things don't last forever, will get better soon, and/or that others have experienced similar struggles.
    \item "Neutralizing": Challenge the negative or catastrophic possibilities and reframe it with a neutral possibility.
    \item "Optimism": Focus and be thankful for the positive aspects of the current situation.
    \item "Self-Affirmation": Say that the character can overcome the challenging event because of their strengths or values.
\end{itemize}

\break

Unhelpful Thinking Pattern Definitions: 
\begin{itemize}[nosep, leftmargin=*]
    \item "Catastrophizing": by giving greater weight to the worst possible outcome.
    \item "Discounting the positive": experiences by insisting that they “don’t count".
    \item "Overgeneralization": making faulty generalizations from insufficient evidence,
    \item "Personalization": assigning a disproportionate amount of personal blame to oneself.
    \item "Black-and-white or polarized thinking / All or nothing thinking": viewing things as either good or bad and nothing in-between.
    \item "Mental filtering": occurs when an individual dwells only on the negative details of a situation.
    \item "Jumping to conclusions: mind reading": inferring a person‘s probable (usually negative) thoughts from their behavior.
    \item "Jumping to conclusions: Fortune-telling": predicting outcomes (usually negative) of events.
    \item "Should statements": a person demands particular behaviors regardless of the realistic circumstances.
    \item "Labeling and mislabeling": attributing a person’s actions to their character rather than the situation.
    \item "None": the thought does not contain any unhelpful pattern / is nonsensical / does not align with the persona. 
\end{itemize}

\subsection{Task 1 Example Prompt (Zeroshot)}
\label{app:task1}
\begin{tcolorbox}
\textbf{You will be given a persona and an unhelpful thought conditioned on the persona. Your goal is to identify the unhelpful thinking pattern that the unhelpful thought falls into.} \vspace{0.5mm} \\

The unhelpful thinking patterns are defined as: \textit{Pattern Definitions}. \vspace{0.5mm} \\

Given a persona and an unhelpful thought, please identify the most appropriate unhelpful thinking pattern. In your response, include only the identified unhelpful thinking pattern from the categories above. \vspace{0.5mm} \\

Persona: \textit{Persona} \vspace{0.5mm} \\
Unhelpful Thought: \textit{Thought} \vspace{0.5mm} \\
Unhelpful thinking pattern:
\end{tcolorbox}

\subsection{Task 2 Example Prompt (Zeroshot)}
\label{app:task2}
\begin{tcolorbox}
\textbf{You will be given a persona and an unhelpful thinking pattern. Your goal is to generate an unhelpful thought that matches the given thinking pattern and the persona.} \vspace{0.5mm} \\

The unhelpful thinking patterns are defined as: \textit{Pattern Definitions}. \vspace{0.5mm} \\

Given a persona and an unhelpful thinking pattern, generate a corresponding unhelpful thought. Contain only the generated unhelpful thought in your response. \vspace{0.5mm} \\

Persona: \textit{Persona} \vspace{0.5mm} \\
Unhelpful thinking pattern: \textit{Pattern} \vspace{0.5mm} \\
Unhelpful thought:
\end{tcolorbox}

\subsection{Task 3 Example Prompt (Zeroshot)}
\label{app:task3}
\begin{tcolorbox}
\textbf{You will be given a persona, an unhelpful thought conditioned on the persona, and the unhelpful thinking pattern it falls into. Your goal is to reframe the unhelpful thought such that it aligns with the persona and context but does not contain the unhelpful thinking pattern.} \vspace{0.5mm} \\

The unhelpful thinking patterns are defined as: \textit{Pattern Definitions}. \vspace{0.5mm} \\

Given a persona, an unhelpful thought, and the unhelpful thinking pattern, please generate a reframed thought. Contain only the reframed thought in your response. \vspace{0.5mm} \\

Persona: \textit{Persona} \vspace{0.5mm} \\
Unhelpful Thought: \textit{Thought} \vspace{0.5mm} \\
Unhelpful thinking pattern: \textit{Pattern} \vspace{0.5mm} \\
Reframing Strategy: \textit{Strategy} \vspace{0.5mm} \\
Reframed Thought:
\end{tcolorbox}

\subsection{Task 4 Example Prompt (Zeroshot)}
\label{app:task4}
\begin{tcolorbox}
\textbf{You will be given a persona, an unhelpful thought conditioned on the persona, the unhelpful thinking pattern that the unhelpful thought falls into, and the reframing strategy used to reframe the thought. Your goal is to reframe the unhelpful thought to be aligned with the reframing strategy while still being aligned with the persona and the context of the unhelpful thought, but without containing the unhelpful pattern.} \vspace{0.5mm} \\

The reframing strategies are defined as: \textit{Strategy Definitions}. \vspace{0.5mm} \\
The unhelpful thinking patterns are defined as: \textit{Pattern Definitions}. \vspace{0.5mm} \\

Given an example of a persona, an unhelpful thought, the unhelpful thinking pattern, and the reframing strategy used, please generate a reframed thought. Contain only the reframed thought in your response. \vspace{0.5mm} \\

Persona: \textit{Persona} \vspace{0.5mm} \\
Unhelpful Thought: \textit{Thought} \vspace{0.5mm} \\
Unhelpful thinking pattern: \textit{Pattern} \vspace{0.5mm} \\
Reframing Strategy: \textit{Strategy} \vspace{0.5mm} \\
Reframed Thought:
\end{tcolorbox}

\end{document}